\documentclass[letterpaper, 10 pt, conference]{ieeeconf}  

\IEEEoverridecommandlockouts                              

\usepackage{cite}
\usepackage{amsmath,amssymb,amsfonts}
\usepackage{graphicx}
\usepackage{textcomp}
\usepackage{color}
\usepackage{hyperref}
\usepackage[noabbrev]{cleveref}
\usepackage{multirow}
\usepackage{tabularray}
\usepackage[T1]{fontenc}
\usepackage{algorithm} 
\usepackage[noend]{algpseudocode}
\usepackage{fancyhdr}
\usepackage{booktabs}
\usepackage{xcolor}
\usepackage{booktabs}
\usepackage{placeins}

\title{\LARGE \bf
Differentiable Mesh State Estimation via Factor Graph \\Inference for Deformable Object Reconstruction}

\author{Lidia Al-Zogbi$^{1*}$, Fangjie Li$^{2*}$, Samuel Tobin$^{3}$, James Ferguson$^{4}$, Nithesh Kumar$^{4}$, Alejandro Chara$^{5}$, \\Kuan-I Chung$^{2}$, Mingxing Rao$^{2}$, Ayberk Acar$^{2}$, Susheela Sharma Stern$^{4}$, Robert Webster$^{4}$, \\Daniel Moyer$^{2}$, Alan Kuntz$^{4}$, Caleb Rucker$^{3}$, Tucker Hermans$^{6,7}$ and Jie Ying Wu$^{2}$
\thanks{Research reported in this publication was supported by the Advanced Research Projects Agency for Health (ARPA-H) under Award Number D24AC00415-00. The award provided 100\% of total costs with an award total of up to \$11,935,038. The content is solely the responsibility of the authors and does not necessarily represent the official views of ARPA-H.}
\thanks{$^{1}$Department of Electrical and Computer Engineering, Tufts University, Medford, MA 02155, USA}
\thanks{$^{2}$Department of Computer Science, Vanderbilt University, Nashville, TN 37235, USA}
\thanks{$^{3}$Department of Mechanical, Aerospace \& Biomedical Engineering, University of Tennessee, Knoxville, TN 37996, USA}%
\thanks{$^{4}$Department of Mechanical Engineering, Vanderbilt University, Nashville, TN 37235, USA}%
\thanks{$^{5}$Vanderbilt University Medical Center, Nashville, TN 37232, USA}%
\thanks{$^{6}$Robotics Center and Kahlert School of Computing, University of Utah, Salt Lake City, UT 84112, USA.}%
\thanks{$^{7}$NVIDIA, Seattle, WA 98105, USA.}%
\thanks{Contact: jieying.wu@vanderbilt.edu}
\thanks{$^{*}$Equal contributions}
}

\begin{document}

\maketitle

\begin{abstract}
Estimating deformable object states remains a fundamental challenge in robotics and simulation. We propose a novel factor graph-based framework for probabilistic mesh state estimation of deformable objects. The method directly updates a tetrahedral mesh---a rich and physically-grounded representation of an environment---by combining physics priors, noisy sensor measurements, and temporal smoothness constraints within a unified probabilistic formulation. The estimation problem is posed as a nonlinear least-squares optimization and solved using Levenberg–Marquardt. \emph{Ex vivo} central-airway obstruction experiments and simulations on deforming cube models demonstrate reliable and accurate reconstruction under both rigid motion and deformation, highlighting the potential of this probabilistic approach for principled, measurement-driven mesh state estimation in deformable object reconstruction.
\end{abstract}

\section{Introduction}
Updating time-varying tetrahedral meshes from point cloud observations is a fundamental inverse problem at the intersection of computational geometry, state estimation, and robotics, enabling compliant manipulation~\cite{yin2021modeling}, predictive simulation~\cite{el2024hyperu}, and real-time digital twinning~\cite{koo2025comprehensive}. Challenges include high-dimensionality of mesh states; partial observability; nonlinear measurement operators that induce non-convex objectives; and the trade-off between measurement fidelity and physics-based regularization.

Existing approaches address subsets of these challenges but lack unified solutions. Kalman-based methods provide principled uncertainty propagation but rely on local linearization that breaks under large deformations and can scale quadratically with state size~\cite{piga2021differentiable, dambreville2006tracking}. Online filters process measurements incrementally but cannot exploit global temporal structure or revise past estimates to improve future predictions. Physics-based variational methods enforce material consistency but require accurate constitutive parameters, are poorly conditioned under measurement noise, and struggle with real-time use~\cite{arriola2020modeling}. Data-driven approaches offer tractability but provide no guarantees of physical plausibility or generalization beyond training~\cite{gao2020learning}.

We propose a factor graph formulation~\cite{loeliger2004introduction, dellaert2012factor} that addresses these limitations through joint spatiotemporal optimization. Our approach treats mesh estimation as a nonlinear least-squares problem where states represent vertex positions, and factors encode physics-based priors, measurement likelihoods, and temporal consistency. This representation enables systematic multi-modal data fusion, and computational efficiency through sparse matrix structures inherent in mesh connectivity.


\section{Related Work}
Factor-graph state estimation provides a natural probabilistic foundation for representing robotic perception and reconstruction problems as structured inference. In this formulation, the posterior distribution over the unknown state is expressed as a product of local factors, where each factor encodes a measurement, prior, motion constraint, or physical relationship among a subset of variables. This representation has been widely adopted in robotics for simultaneous localization and mapping, visual-inertial estimation, and smoothing because it exposes the sparse structure of the estimation problem and enables efficient nonlinear least-squares optimization~\cite{dellaert2017factor,dellaert2012gtsam}. While traditional applications often estimate robot poses, landmarks, or calibration parameters, the same formulation can be extended to richer environment states. In our work, the estimated variable is not a low-dimensional pose, but an explicit mesh representation of the deformable object or environment. This allows the posterior over the mesh state to be constructed from heterogeneous factors, including geometric measurement terms, temporal consistency, boundary conditions, and mechanics-informed priors.

Differentiable geometric optimization has similarly become an important tool for estimating 3D structure from observations. Many geometric estimation problems can be written as optimization problems over residuals defined on points, surfaces, images, meshes, or implicit fields, and can be solved using Gauss-Newton, Levenberg-Marquardt, or gradient-based optimization. General-purpose solvers such as Ceres support nonlinear least-squares formulations with automatic differentiation, while differentiable 3D libraries such as PyTorch3D provide differentiable operators for meshes, point clouds, rendering, and geometric losses~\cite{agarwal2023ceres,ravi2020pytorch3d}. These tools have enabled analysis-by-synthesis and gradient-based reconstruction pipelines in which the geometric state is updated directly to reduce observation error. Our formulation builds on this perspective by defining residuals that are differentiable with respect to the mesh state itself. As a result, point-to-surface measurements, temporal regularization terms, and stiffness-based priors can all contribute gradients to a common optimization problem, enabling mesh reconstruction to be treated as differentiable inference rather than as a purely procedural registration step.

Deformable environment modeling addresses the challenge of estimating objects or scenes whose geometry changes over time. Classical approaches include non-rigid registration, deformation graphs, and surface regularization methods that recover a deformation field or deformed geometry from sparse or dense observations. Coherent Point Drift formulates non-rigid point-set registration probabilistically by aligning one point set to another while enforcing coherent motion of the transformed points~\cite{myronenko2010cpd}. Embedded deformation represents shape change using a sparse graph of local transformations that can deform complex geometry through direct manipulation~\cite{sumner2007embedded}, while as-rigid-as-possible surface modeling regularizes deformation by encouraging local transformations to remain close to rigid~\cite{sorkine2007arap}. Dense reconstruction systems such as DynamicFusion further demonstrate that non-rigid scene geometry can be reconstructed and tracked in real time from RGB-D observations by jointly estimating geometry and a deformation field~\cite{newcombe2015dynamicfusion}. These methods are highly relevant to deformable reconstruction, but they often treat deformation primarily as geometric alignment, tracking, or warping. In contrast, our work frames deformable object reconstruction as state estimation over an explicit mesh-valued environment state, allowing measurements and priors to be expressed as probabilistic factors within a unified inference problem.

Physics-informed estimation provides another important foundation for deformable reconstruction because purely geometric alignment can produce deformations that match observations locally but are physically implausible globally. Deformable modeling has long incorporated mechanical principles such as elasticity, stiffness, internal forces, damping, and constraints to describe how non-rigid bodies respond to forces and boundary conditions~\cite{baraff1998large}. In computer graphics and simulation, such models have been used to produce stable and realistic deformable motion, while in robotics and perception they provide priors that constrain the space of admissible environment states. For mesh-based reconstruction, stiffness or mechanics-inspired regularization can encode the intuition that neighboring vertices, material regions, or constrained boundaries should deform in a physically consistent manner. Our approach incorporates this idea by using stiffness-informed factors as probabilistic priors over the mesh state. Rather than relying only on geometric correspondence, the estimator combines observation likelihoods with mechanical regularization and temporal consistency, producing a posterior estimate that is both measurement-driven and physically constrained. This is especially important for deformable object reconstruction during manipulation and cutting, where observations may be sparse, partial, noisy, or locally ambiguous.

\section{Factor Graph Formulation}
\subsection{Baseline Deformation Model}
\noindent\textbf{State Representation}: We represent the deformable object as a tetrahedral mesh with $N$ vertices, where the state variables are the 3D coordinates of all mesh vertices. We assume the mesh connectivity is given a priori and does not change over time.
At each time step $t$, the mesh state is  $\mathbf{X}_t =\big[\mathbf{x}_{t,1}^\top,\dots,\mathbf{x}_{t,N}^\top\big]^\top \in \mathbb{R}^{3N}$
, with each vertex position $\mathbf{x}_{t,i} \in \mathbb{R}^3$.


\noindent\textbf{Factor Graph Formulation}: In a factor graph, each factor encodes a probabilistic constraint on selected state variables, represented by a residual function and weighted by its information matrix (the inverse covariance). We consider three principal factor types:

\begin{itemize} 
    \item \textit{Physics Prior Factor}: 
    Enforces physically plausible nodal displacements by using the finite element stiffness matrix of the mesh as the information matrix $\Lambda^{\text{phys}}_t=\mathbf{K}_t(\mathbf{X}_t)$. Note the dependence of $\mathbf{K}_t$ on the current state $\mathbf{X}_t$.

    We model temporal evolution of the mesh as a probabilistic transition between consecutive states \(P(\mathbf{X}_{t}|\mathbf{X}_{t-1}) = \mathcal{N}(\mathbf{X}_{t} | \hat{\mathbf{X}}_{t}, \mathbf{K}_t^{-1})\). The residual for the prior factor is:
    \begin{equation}
    \mathbf{r}^{\text{phys}}_t(\mathbf{X}_t) =  \mathbf{\hat{X}}_{t} - \mathbf{X}_{t},
    \end{equation}
    where a deformable simulator~\cite{tobin2025efficient} predicts $\mathbf{\hat{X}}_{t}$ from $\mathbf{X}_{t-1}$.

    \item \textit{Point-to-Surface Measurement Factor}:
Penalizes error between measurements and mesh surface. We associate every measurement point $\mathbf{m}_{t,k}\in\mathbb{R}^3$ with its closest point on the mesh surface at time step $t$, where $k\in[1, K]$ and $K$ is the number of measurements. 
Let $\mathcal{T}$ denote the set of surface triangles with vertex indices $(a,b,c)$. 
For each measurement $\mathbf{m}_{t,k}$ we identify the surface triangle whose projection is closest to the measurement point:
\begin{equation}
    a^\star,b^\star,c^\star
    = \operatorname*{argmin}_{(a,b,c)\in\mathcal{T}} 
    \; \big\| \Pi_{\triangle(a,b,c)}(\mathbf{m}_{t,k}) - \mathbf{m}_{t,k} \big\|^2,
\end{equation}
where $\Pi_{\triangle(a,b,c)}(\cdot)$ is the closest-point projection operator onto the triangle with vertices
$(\mathbf{x}_{t,a},\mathbf{x}_{t,b},\mathbf{x}_{t,c})$. This projection operator admits a closed-form solution in barycentric coordinates $(\alpha,\beta,\gamma)$, 
with $\alpha+\beta+\gamma=1,\;\alpha,\beta,\gamma\ge 0$ obtained by standard point-to-triangle projection algorithms~\cite{heidrich2005computing}. The \(k\)-th measurement residual is:
\begin{equation}
    \mathbf{r}^{\text{meas}}_{t,k} = 
    \big(\alpha \mathbf{x}_{t,a^\star} + \beta \mathbf{x}_{t,b^\star} + \gamma \mathbf{x}_{t,c^\star}\big) - \mathbf{m}_{t,k},
\end{equation}
which can be used to form the concatenated residual vector across all measurements as $\mathbf{r}_t^{\text{meas}} = \big[\mathbf{r}^{\text{meas}\top}_{t,1}, ..., \mathbf{r}^{\text{meas}\top}_{t,K}]^\top$. 
$\Lambda_t^{\text{meas}}$ denotes the information matrix encoding the sensor noise model for all measurements at time $t$.

When the closest triangle $(a^\star,b^\star,c^\star)$ remains fixed, 
the projection is differentiable with respect to its three vertex positions. 

    \item \textit{Temporal Smoothness Factor}: Connects consecutive states to encourage temporal smoothness of the mesh. The residual is thus:
    \begin{equation}
\mathbf{r}^{\text{temp}}_t(\mathbf{X}_t) =  \mathbf{X}_{t} - \mathbf{X}_{t-1},
    \end{equation} penalizing large inter-frame displacements. The associated information matrix $\Lambda^{\text{temp}}_t$ controls the strength of this smoothness prior, and is chosen as a scaled identity matrix assuming isotropic uncertainty across vertices.
\end{itemize}

\subsection{Optimization}
The maximum a posteriori (MAP) estimation problem can be reformulated as a nonlinear least-squares objective~\cite{dellaert2012factor} over all mesh states $\mathbf{X} = \{\mathbf{X}_0, \ldots, \mathbf{X}_t\}$:
\begin{equation}
    \mathbf{X}^\star = \operatorname*{argmin}_\mathbf{X} 
     \sum_t\left(\mathcal{L}^{\text{phys}}_t + \mathcal{L}^{\text{meas}}_t + \mathcal{L}^{\text{temp}}_t\right).
\end{equation}
As time advances, new states $\mathbf{X}_{t+1}$ and their associated factors are appended to the graph, so the optimization dimension grows while preserving past information. Each factor $i$ contributes a quadratic cost
$\mathcal{L}_i = \tfrac{1}{2}\, \mathbf{r}_i(\mathbf{X})^\top \Lambda_i \mathbf{r}_i(\mathbf{X})$, where $\mathbf{r}_i(\mathbf{X})$ is the residual vector and $\Lambda_i$ the associated information matrix. The loss function is solved iteratively using the damped Levenberg–Marquardt~\cite{more2006levenberg} algorithm.


This work represents a prototype implemented in Python~3.12, using the PyTorch library~\cite{paszke2019pytorch} for automatic differentiation and accelerated numerical computations.

\section{Experiments}
\subsection{Simulation-Based}
\subsubsection{Ablation Studies}
We performed an ablation study to evaluate the contribution of each factor type in the proposed mesh state estimation framework. The study was conducted using forward-simulated deforming-cube experiments with unit side length, including: (a) rigid-body translation, (b) deformation with an SPD stiffness prior, and (c) deformation with a non-SPD stiffness prior. For each case, the tetrahedral mesh state was reconstructed from synthetic surface measurements sampled from the deformed mesh and perturbed with Gaussian noise. The forward-simulation vertices were used as ground truth, and reconstruction accuracy was quantified using the vertex RMSE between the estimated and ground-truth mesh states. We used a cube with 1 meter side length for the simulation.

We evaluated five factor configurations: temporal smoothness only, measurement only, temporal and measurement factors, simulation and measurement factors, and the full model combining simulation, measurement, and temporal factors. The temporal factor penalizes inter-frame vertex displacement, the measurement factor penalizes point-to-surface discrepancy between noisy observations and the current mesh surface, and the simulation factor incorporates physics-based information from the forward simulator. The physics prior information matrix was obtained from the simulation stiffness matrix, the measurement factors used identity information matrices assuming independent observations, and the temporal smoothness factors used scaled identity matrices.

Errors were computed over three vertex sets: all vertices, all external vertices, and unobserved external vertices. The last metric is particularly important because it evaluates whether the estimator can infer the state of surface regions that are not directly constrained by measurements. Across the simulated cases, the overall RMSE increased for the deformable experiments, particularly when using the non-SPD prior, likely due to mesh warping and the simplified identity measurement covariance. Nevertheless, the solution remained stable across selected time steps, with relatively small reconstruction error even in poorly conditioned settings.

\subsection{Ex Vivo Studies}
\subsubsection{Clinical Motivation}
Central airway obstruction (CAO) is a disorder of increasing prevalence and is associated with significant morbidity and mortality~\cite{ernst2004central}. Therapeutic tools such as laser ablation and electrocautery, delivered through rigid or flexible bronchoscopes, can enable removal of obstructing tumor tissue and restoration of airway patency~\cite{ernst2004central}. However, bronchoscopic intervention for CAO remains technically challenging: the operative workspace is narrow, the anatomy is deformable, visibility may be partial, and complications such as bleeding, airway perforation, hypoxia, or loss of airway control can be fatal~\cite{stahl2015complications,gafford2020concentric}. These challenges motivate robotic systems that can maintain an accurate estimate of airway and tumor geometry during tissue manipulation and cutting.

\subsubsection{Experimental Setup}
In this work, we use an \emph{ex vivo} CAO model to evaluate whether the proposed factor graph formulation can estimate deformable tissue state from partial observations during robot-assisted intervention. The goal is not only to reconstruct visible surface geometry, but also to infer the volumetric deformation of the CAO model so that unobserved regions remain registered to the robotic system throughout deformation.

\begin{figure}[t]
    \centering
    \includegraphics[width=\columnwidth]{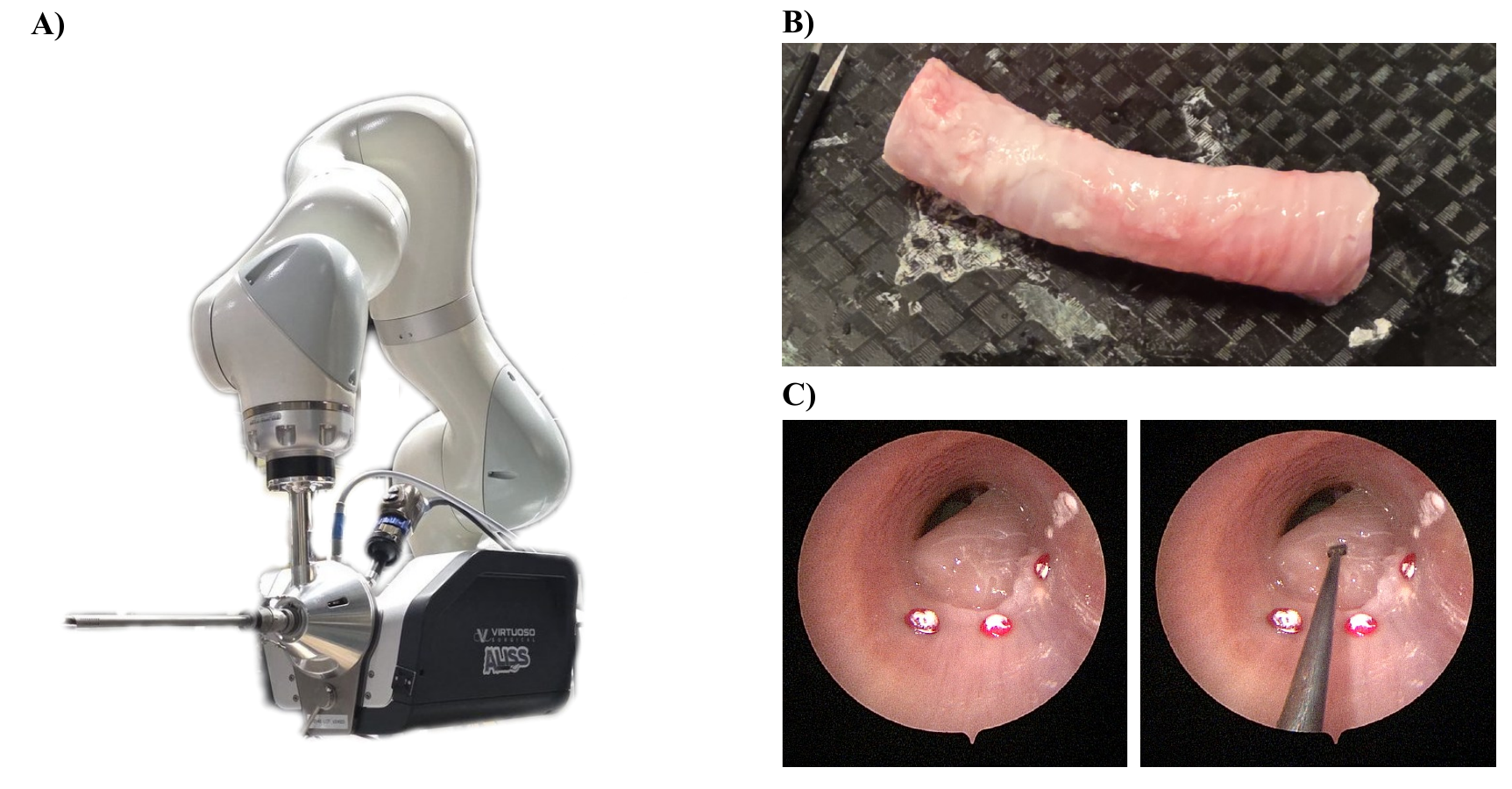}
    \caption{A) The Virtuoso robot system used for the \emph{ex vivo} experiment. B) The sheep trachea used as CAO phantom. C) Endoscopic view of the CAO tumor, with and without tool interaction. There are additional red fiducials in the phantoms, although not used in this experiment.}
    \label{fig:setup}
\end{figure}

We performed the \emph{ex vivo} study on a CAO phantom using a Virtuoso robotic system (Virtuoso Surgical, Inc., Nashville, TN, USA), as outlined in Fig.~\ref{fig:setup}. Following our previous work~\cite{acar2025monocular, li2025automated}, we use sheep trachea to mimic human anatomy. We cut and insert a small piece of chicken breast into the trachea to mimic tumor that causes around 50\% occlusion in the trachea. It is placed inside the airway through small incisions, secured with super glue.

During the experiment, we used the method outlined in~\cite{acar2026perseus} to register the segmented pre-operative CT into the robot coordinate frame, which is used to initialize the tissue simulation method. During the experiment, we manually deformed the tumor with the robotic tool. We then used a xCAT CT system (Xoran Technologies LLC, Ann Arbor, MI, USA) to obtain a snapshot ground truth shape of the deformed anatomy. We also capture the relevant input factors to the factor graph algorithm at 1 Hz. We then run the factor graph algorithm for ten time steps, and use the final step output for evaluation. These intra-operative CT volumes are registered to the pre-operative CT using fiducial markers rigidly attached to the exterior of the CAO phantom. We had 3 CAO phantom models, with 3 pushes on each phantom, giving 9 experimental trials in total. They were excluded from the MDE training data.

\subsubsection{Factors}
The \emph{ex vivo} factor graph used the same volumetric mesh state representation as the simulation studies, with $\mathbf{X}_t$ denoting the tetrahedral CAO mesh in the robot base frame at time $t$. Similar to the simulation case, two primary factors were used: an XPBD-based simulation factor and a metric monocular-depth-estimation (MDE) point-cloud measurement factor.


The MDE factor incorporates a point cloud reconstructed from an MDE network model, constituting the point-to-surface measurement factor above. It is  built on top of a non-metric depth estimator, DepthAnything-v2~\cite{yang2024depth}. This depth estimator is in turn built on a DINOv2~\cite{oquab2023dinov2} backbone and teacher model, which is a generic natural vision foundation model. DINOv2 is a Vision Transformer~\cite{dosovitskiy2020image} architecture model with contrastive self-supervised pre-training tasks; DepthAnything-v2 supplements this with additional synthetic scene training on depth specific tasks and a Teacher-Student distillation. Our present metric MDE additionally fine-tunes this model in two stages, first across all weights and second across the depth prediction head. For both of these stages we train using 6000 video frames, with supervision in the form of depth information derived from CT registered to the robotic viewport, using the $\ell_2$ Euclidean loss instead of SigLog loss (for relative non-metric depth). 

We further incorporate a segmentation network to segment out the tumor only, to remove non-tumor depth predictions, such as that of the tool and the trachea. The segmentation network consists of (1) a frozen SAM2~\cite{ravi2024sam2} Hiera image encoder with lightweight adapters for efficient fine-tuning, (2) multi-branch convolutional feature modules that refine and unify multi-scale features extracted by the encoder, and (3) a U-Net-style~\cite{ronneberger2015unet} decoder that progressively fuses the refined multi-scale features through skip connections to generate the final segmentation. During training, only the adapters, the feature refinement modules, and the decoder are trainable, while the pretrained SAM2 encoder remains frozen.

The uncertainty of the MDE factor is set empirically based on its error evaluation.




\subsubsection{Quantitative Evaluation}

\begin{table}[t]
\caption{Simulation Results (Mean $\pm$ Standard Deviation, mm)}
\label{tab:simulation_ablation}
\centering
\footnotesize
\setlength{\tabcolsep}{3.5pt}
\begin{tabular}{lccc}
\hline
\textbf{Factor configuration}
    & \textbf{All}
    & \textbf{External}
    & \shortstack{\textbf{Unobserved}\\\textbf{external}} \\
\hline
Temporal only
    & $5.5 \pm 3.4$
    & $5.3 \pm 3.3$
    & $5.3 \pm 3.2$ \\
Measurement only
    & $280 \pm 220$
    & $300 \pm 240$
    & $350 \pm 280$ \\
Temporal + measurement
    & $120 \pm 9.0$
    & $98 \pm 9.1$
    & $110 \pm 11$ \\
Simulation + measurement
    & $0.73 \pm 0.22$
    & $0.77 \pm 0.21$
    & $0.67 \pm 0.23$ \\
Full model
    & $0.73 \pm 0.22$
    & $0.76 \pm 0.21$
    & $0.67 \pm 0.23$ \\
\hline
\end{tabular}
\end{table}

\begin{table*}
\centering
\caption{Surface reconstruction errors over the full and unobserved surfaces (mean $\pm$ standard deviation, mm).}
\label{tab:full_unlocalized_surface_error}
\setlength{\tabcolsep}{4pt}
\begin{tabular}{lccc|ccc}
\hline
& \multicolumn{3}{c|}{\textbf{Full Surface}}
& \multicolumn{3}{c}{\textbf{Unobserved Surface}} \\
\textbf{Method}
& \textbf{$d_{CD}$}
& \textbf{$d_{RMSE}$}
& \textbf{$d_{H95}$}
& \textbf{$d_{CD}$}
& \textbf{$d_{RMSE}$}
& \textbf{$d_{H95}$} \\
\hline
Simulation
& $1.97 \pm 0.60$
& $2.77 \pm 0.84$
& $5.28 \pm 1.46$
& $1.92 \pm 0.59$
& $2.57 \pm 0.82$
& $5.09 \pm 1.58$ \\


Factor Graph
& $\mathbf{1.61} \pm 0.67$
& $\mathbf{2.41} \pm 0.96$
& $\mathbf{4.34} \pm 1.63$
& $\mathbf{1.51} \pm 0.65$
& $\mathbf{2.23} \pm 0.72$
& $\mathbf{4.12} \pm 1.40$ \\
\hline
\end{tabular}
\end{table*}

\begin{table}[!t]
\centering
\caption{Surface reconstruction errors within the MDE observable region
(mean $\pm$ standard deviation, mm).}
\label{tab:localized_surface_error}
\setlength{\tabcolsep}{3pt}
\footnotesize
\begin{tabular}{lccc}
\hline
\textbf{Method}
& \textbf{$d_{CD}$}
& \textbf{$d_{RMSE}$}
& \textbf{$d_{H95}$} \\
\hline
Simulation
& $1.84 \pm 0.71$
& $2.57 \pm 1.14$
& $4.81 \pm 1.80$ \\


Factor Graph
& $\mathbf{1.63} \pm 0.78$
& $2.38 \pm 1.26$
& $\mathbf{4.26} \pm 1.76$ \\

MDE
& $2.16 \pm 0.86$
& $\mathbf{1.94} \pm 0.41$
& $7.44 \pm 4.85$ \\
\hline
\end{tabular}
\end{table}

For quantitative evaluation, the optimized mesh was sampled to produce an estimated point cloud
$\hat{\mathcal{P}}_t=\{\hat{\mathbf{p}}_{t,i}\}_{i=1}^{N_t}$, which was compared against a reference point cloud
$\mathcal{P}^{\mathrm{ref}}_t=\{\mathbf{p}^{\mathrm{ref}}_{t,j}\}_{j=1}^{M_t}$ in the robot base frame. We first computed the one-sided nearest-neighbor error from the estimated reconstruction to the reference:
\begin{equation}
d_{\hat{\mathcal{P}}\rightarrow\mathcal{P}^{\mathrm{ref}}}
=
\frac{1}{N_t}
\sum_{i=1}^{N_t}
\min_{j}
\left\|
\hat{\mathbf{p}}_{t,i}
-
\mathbf{p}^{\mathrm{ref}}_{t,j}
\right\|_2 .
\end{equation}
The reverse error was similarly computed as:
\begin{equation}
d_{\mathcal{P}^{\mathrm{ref}}\rightarrow\hat{\mathcal{P}}}
=
\frac{1}{M_t}
\sum_{j=1}^{M_t}
\min_{i}
\left\|
\mathbf{p}^{\mathrm{ref}}_{t,j}
-
\hat{\mathbf{p}}_{t,i}
\right\|_2 .
\end{equation}
The symmetric Chamfer distance was then defined as:
\begin{equation}
d_{\mathrm{CD}}
= \frac{1}{2}
\left(d_{\hat{\mathcal{P}}\rightarrow\mathcal{P}^{\mathrm{ref}}}
+
d_{\mathcal{P}^{\mathrm{ref}}\rightarrow\hat{\mathcal{P}}}\right).
\end{equation}

We additionally report the root-mean-square nearest-neighbor error:
\begin{equation}
d_{\mathrm{RMSE}}
=
\sqrt{
\frac{1}{N_t}
\sum_{i=1}^{N_t}
\min_{j}
\left\|
\hat{\mathbf{p}}_{t,i}
-
\mathbf{p}^{\mathrm{ref}}_{t,j}
\right\|_2^2
}.
\end{equation}
To capture worst-case geometric disagreement while reducing sensitivity to isolated outliers, we also report the $95$th-percentile  distance:
\begin{equation}
\begin{aligned}
d_{\mathrm{H95}}
=
\max \bigg(
&
Q_{0.95}
\left(
\left\{
\min_j
\left\|
\hat{\mathbf{p}}_{t,i}
-
\mathbf{p}^{\mathrm{ref}}_{t,j}
\right\|_2
\right\}_{i=1}^{N_t}
\right),
\\
&
Q_{0.95}
\left(
\left\{
\min_i
\left\|
\mathbf{p}^{\mathrm{ref}}_{t,j}
-
\hat{\mathbf{p}}_{t,i}
\right\|_2
\right\}_{j=1}^{M_t}
\right)
\bigg).
\end{aligned}
\end{equation}
where $Q_{0.95}(\cdot)$ denotes the $95$th percentile. Together, these metrics quantify both average reconstruction accuracy and localized geometric disagreement between the estimated and reference point clouds.

As shown in Fig.~\ref{fig:qualitative_result}, we only consider the lesion surface for metric computation. This is because the lesion-trachea interface is considered to be fixed, and hence including it in the measurement will skew the error. Additionally, there is no ground truth correspondence between the sub-surface vertices and the CT scan.

To further analyze the effect of the MDE measurements on the mesh region not observable by MDE, we separate the mesh into 2 regions. We perform ray-tracing from the camera optical center to the sampled MDE point cloud. The intersection of the rays and $\hat{\mathcal{P}}_t$ 
and $\mathcal{P}^{\mathrm{ref}}_t$ defines a proxy subset of the point clouds that would have been directly observable by MDE.

We also compare the outputs from factor graph, against its contributing factors, tissue simulation and MDE measurements.


\section{Results and Discussion}
\subsection{Simulation}
Table~\ref{tab:simulation_ablation} summarizes the contribution of
each factor type to mesh-state reconstruction over the first 15 time
steps of the deformable-cube shear experiment. The measurement-only
configuration produced the largest errors, with a corresponding-vertex
RMSE of $280 \pm 220~\mathrm{mm}$ over all vertices and
$350 \pm 280~\mathrm{mm}$ over unobserved external vertices. Adding
temporal smoothness reduced these errors to $120 \pm 9.0~\mathrm{mm}$
and $110 \pm 11~\mathrm{mm}$, respectively. However, this configuration
remained poorly constrained, and several optimization steps reached the
maximum iteration limit.

The temporal-only configuration produced smaller errors of
$5.5 \pm 3.4~\mathrm{mm}$ over all vertices and
$5.3 \pm 3.2~\mathrm{mm}$ over unobserved external vertices. This result
should be interpreted cautiously: without an absolute factor, the
copy-forward initialization leaves the mesh at its initial state, which
remains close to the ground truth over this relatively short portion of
the deformation trajectory.

\begin{figure}[t]
    \centering
    \includegraphics[width=\columnwidth]{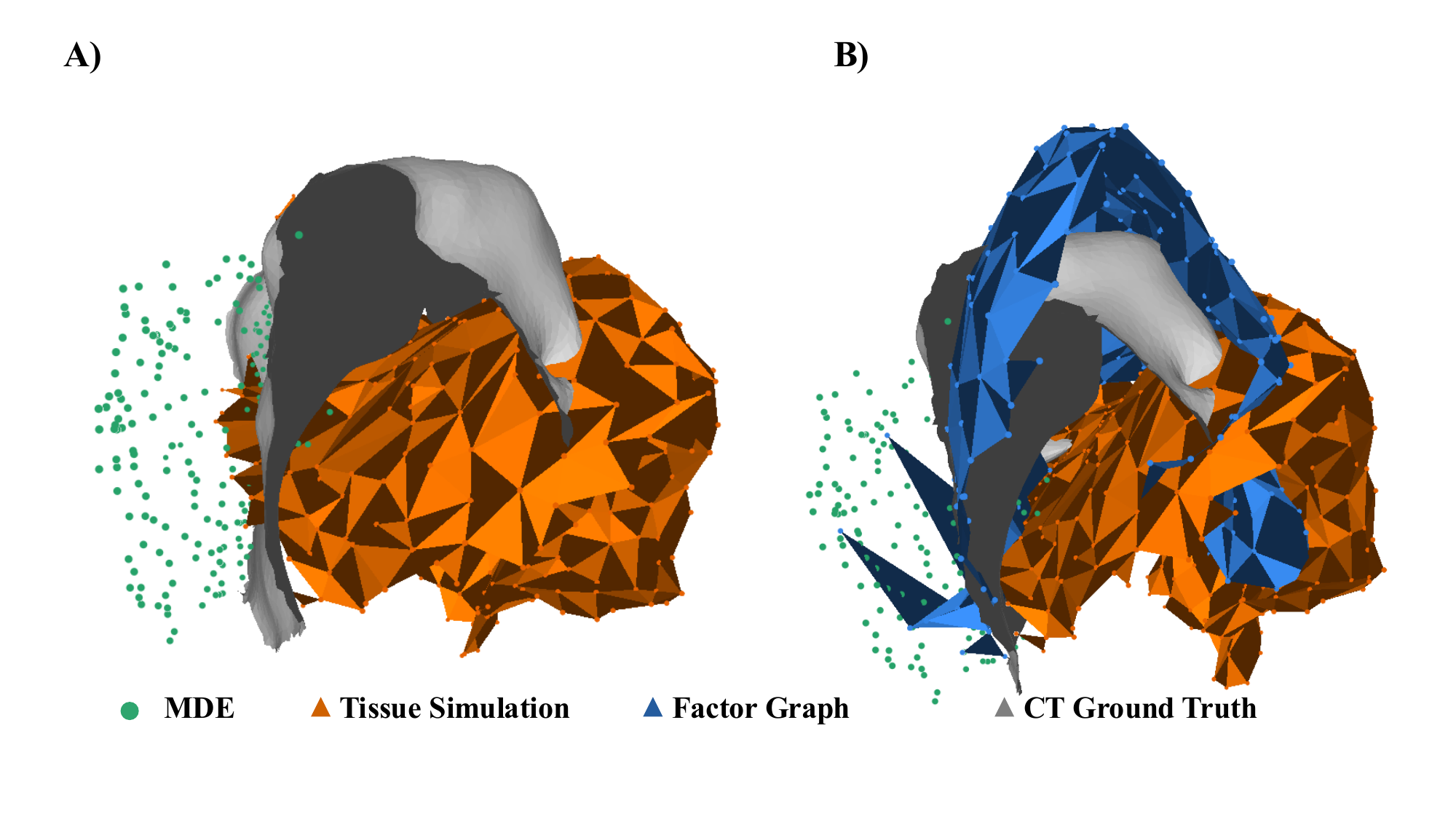}
    \caption{A) The side view of the factors as input to the factor graph, along with the ground truth. B) The factor graph output super-imposed.}
    \label{fig:qualitative_result}
\end{figure}

Incorporating the simulation prior produced the lowest reconstruction
errors. The simulation-and-measurement configuration achieved errors of
$0.73 \pm 0.22~\mathrm{mm}$ over all vertices,
$0.77 \pm 0.21~\mathrm{mm}$ over external vertices, and
$0.67 \pm 0.23~\mathrm{mm}$ over unobserved external vertices. The full
model produced nearly identical errors of
$0.73 \pm 0.22~\mathrm{mm}$, $0.76 \pm 0.21~\mathrm{mm}$, and
$0.67 \pm 0.23~\mathrm{mm}$, respectively. These results indicate that
the physics-based simulation prior is the principal contributor to
accurate reconstruction in this experiment, while the temporal factor
provides no measurable improvement once the simulation prior is
present. This effect is illustrated in
Fig.~\ref{fig:qualitative_result}.


\subsection{Ex Vivo Studies}

Table~\ref{tab:full_unlocalized_surface_error} describes the alignment of the estimated tumor surfaces by factor graph and tissue simulation against the ground truth tumor state. The full factor graph achieves a lower Chamfer distance than tissue simulation alone. Notably, this improvement is also observed in regions not directly observed by the MDE measurement. This suggests that the tissue-mechanics prior encoded by the stiffness matrix can help propagate information from observed regions to constrain the unobserved mesh state within the factor graph framework.

Table~\ref{tab:localized_surface_error} outlines the result for the MDE observable region. Here, factor graph displays the lowest $d_{CD}$ and $d_{\mathrm{H95}}$, MDE has the lowest $d_{\mathrm{RMSE}}$. This suggests that MDE has higher local precision, whereas the lower two-way $d_{CD}$ and $d_{H95}$ of the factor graph indicate better coverage of the tumor surface. This may be attributed to the factor graph explicitly modeling the complete tumor mesh. Near regions viewed at grazing angles, the MDE point cloud becomes increasingly sparse over the physical surface, and small image-space or pose misalignments can correspond to larger surface displacements. Consequently, the ground-truth-to-MDE component of the bidirectional metrics can be disproportionately affected in these regions.

Taken together, these results suggest the complementary roles of the two information sources within the factor graph framework. MDE provides locally precise but spatially limited measurements, while tissue simulation provides a global tissue-mechanics prior that constrains the overall deformation state. By jointly incorporating these sources, the factor graph can leverage accurate local observations while propagating their information through the mechanical prior to improve estimation of the complete tissue state, including regions without direct MDE observations.

\FloatBarrier

\section{Conclusions}
Preliminary experiments show that the proposed factor graph formulation closely tracks ground-truth deformations in simulation and \emph{ex vivo} CAO experiments, highlighting its promise for mesh-based state estimation of deformable objects. Future work will incorporate sensor noise models beyond identity-weighted information matrices, extend validation to additional physical experiments, and address complex cases such as mesh topology changes during cutting~\cite{heiden2021disect}. By unifying measurements and physics within a probabilistic framework, this approach has the potential to transform how robots perceive, predict, and interact with deformable environments.

\newpage
\bibliography{bibliography}
\bibliographystyle{ieeetr}

\end{document}